\documentclass{bmvc2k}
\usepackage{amsmath}
\usepackage{underscore}
\usepackage{multirow}
\usepackage{float}
\usepackage{graphicx}
\usepackage{subfigure}
\usepackage{xspace}
\usepackage{pdfpages}
\usepackage{setspace}
\usepackage{float}
\usepackage{geometry}
\usepackage[ruled,linesnumbered]{algorithm2e}

\title{Copy and Paste method based on Pose for Re-identification}

\addauthor{Cheng Yang}{}{1}

\addinstitution{
Huazhong University of \\
Science and Technology. \\
Wuhan, 430074, China.
}

\runninghead{Copy and Paste method based on Pose for Re-identification}{}

\begin{document}

\maketitle

\begin{abstract}
The aim of re-identification is to match objects in surveillance cameras with different viewpoints. Although ReID is developing at a considerably rapid pace, there is currently no processing method for the ReID task in multiple scenarios. However, such processing method is required in real life scenarios, such as those involving security. In the present study, a new ReID scenario was explored, which differs in terms of perspective, background, and pose(walking or cycling). Obviously, ordinary ReID processing methods cannot effectively handle such a scenario, with the introduction of image datasets being the optimal solution, in addition to being considerably expensive.

To solve the aforementioned problem, a simple and effective method to generate images in several new scenarios was proposed, which is names the Copy and Paste method based on Pose(\textit{CPP}). The \textit{CPP} method is based on key point detection, using copy as paste, to composite a new semantic image dataset in two different semantic image datasets. As an example, pedestrains and bicycles can be used to generate several images that show the same person riding on different bicycles. The \textit{CPP} method is suitable for ReID tasks in new scenarios and outperforms the traditional methods when applied to the original datasets in original ReID tasks. To be specific, the \textit{CPP} method can also perform better in terms of generalization for third-party public dataset. The Code and datasets composited by the \textit{CPP} method will be available in the future.
\end{abstract}

\section{Introduction}
\label{sec:intro}
The aim of re-identification(ReID) is to match images of an object across different cameras, being widely used in video surveillance, security, smart cities and others. Various methods have recently been proposed for ReID, but most of methods focus on holistic images and neglect multiple scenarios cases, which may be more challenging. For example, a scenario in which a thief is stealing in a community and the gets on a bicycle and escapes is more complex. In analysing from visual perspective, a number of factors are involved in such a scenario, including the change of background,  the irregular occlusion of the bicycle, and the change of the character's poses. Thus. matching the image of the person with multiple scenes and poses in which they are located is necessary, which is known as the multiple scenarios ReID problem.

Compared with matching people with normal scenarios, ReID with multiple is more challenging due to the following reasons: (1) with the background regions, the multiple scenarios have more complex background differences; (2) with the occluded regions, the multiple scenarios have more complex foreground information, but in the tasks of ReID with multiple scenarios, matching people may be misleading\cite{r1, r2}, one example being that bicycle of a cyclist is not valid feature information for person ReID tasks; and (3) with regard to pose, a person has different poses in different scenarios, which is one of the main issues involved in person ReID tasks.

Neural Networks for ReID are developing rapidly, with specific datasets and evaluation for specific scenarios. For instance, there are Occluded-DukeMTMC\cite{r3, r4} and Occluded-REID\cite{r5} datasets for occluded scenes. In a similar vein, for the proposed multiple scenarios in the present study, the optimal solution was to construct corresponding datasets, but no such datasets existed for the present research. As such, a simple and effective method of constructing datasets was proposed, involving the merging of two datasets with different semantic content into one dataset with new semantic content. Said dataset can be used for ReID tasks in multiple scenarios, and the method is named the Copy and Paste method based on Pose (\textit{CPP}), which includes pose detection, affine transformation, and copy and paste. Different from previous research on network design, the aim of the present study was to build new application scenarios for ReID tasks, and datasets suitable for the new scenarios. 

The main contributions of the present study paper are summarized as follows: (1) new scenarios were proposed for ReID tasks, and a method for the construction of datasets therefor was provided; (2)  the new dataset generated by the \textit{CPP} method was used, which could be applied to the corresponding new scenarios, and the new dataset could offer better results than the original ReID tasks of the original dataset, with the same being true for the third-party public ReID dataset; and (3) the code and datasets composited by the \textit{CPP} method will be available in the feature.

\section{Related Work}
\label{sec:related}
\textbf{Pedestrian ReID.}  Person re-identification addresses the problem of matching pedestrian images across disjointed cameras. The primary challenges involved in person ReID are the large intra-class and small inter-class variations caused by different backgrounds, viewpoints, poses, and other factors. The existing research in the present field has predominantly focused on the development of loss functions\cite{r6, r7} or network architectures\cite{r11, r12, r13, r14}. In terms of the former, a series of deep metric learning methods\cite{r8, r9, r10, r1} have been proposed to enhance intra-class compactness and inter-class separability in the manifold. As for the network of architectures, some of them treat it as a partial feature learning task and pay much attention to more powerful network structures to better discover, align, and extract the body parts features\cite{r16, r18, r19}. The above works is based on current mainstream public datasets, such as Market1501\cite{r20}, DukeMTMC-reID\cite{r21}, MSMT17\cite{r22}, etc. They are all pedestrain datasets and their action are very similar, and it's not as complicated in multiple scenarios.

\textbf{Vihicle ReID.}  The aim of vehicle ReID is to match vehicles in surveillance cameras with different viewpoints, which is of significant value in public security and intelligent transportation. The major challenge is the viewpoint variation problem\cite{r23, r28, r25, r26}. Vehicle re-identification benefits from the progresses of person re-identification [4, 5, 6]. But vehicle re-identification still has many challenges, including intra-class variations for vehicle-model, views, occlusion, motion blur\cite{r27}. The follow-up works focus on the discriminative representation learning\cite{r28, r29} as well as mining the structure information\cite{r30} For example, some methods\cite{r31, r32} propose to leverage the multiscale information within deeply-learned models. To mine the fine-grained vehicle structure, some works\cite{r33} further take the keypoints into consideration and apply the structure information to the final feature aggregation part. Since the vehicle is a rigid object, it has a better distinction from the appearance. And it is difficult to discuss with person in ReID tasks.

\textbf{Occlusion ReID.} The aim of occlusion ReID is to match occluded images to holistic images across disjointed cameras. Due to incomplete information and spatial misalignment, such task can be considerably challenging. For occlusion ReID, several methods\cite{r5, r35, r36} use occluded/non-occluded binary classification loss to distinguish the occluded images from holistic images. Another method involves a saliency map being predicted to highlight the discriminative parts, with a teacher-student learning scheme further improving the learned features. A pose guided feature alignment method\cite{r3}  was also proposed to match the local patches of probe and gallery images based on the human semantic key-points. Although the aforementioned solutions have made good progress in solving occlusion problems, due to factors such as shape and color, the occlusions are simpler to handle than multiple scenarios.

\textbf{Cross Module.} Cross-modality ReID overcomes the new challenge of not being able to effectively complete image retrieval of RGB images in certain scenes. The aim of cross-modality ReID is to match queries of one modality against a gallery set of another modality, such as text-image ReID\cite{r38, r39, r40}, RGB-Depth ReID\cite{r41} and RGB-Infrared(RGB-IR) ReID \cite{r42}, the \cite{r43} built the largest SYSU-MM01 dataset for RGB-IR person ReID evaluation. The \cite{r44} advanced a two-stream based model and bi-directional top-ranking loss function for the shared feature embedding. To make the shared features purer, the \cite{r45} suggested a generative adversarial training method for the shared feature learning. More recently, The \cite{r46} constructed a novel GAN model with the joint pixel-level and feature-level constraint, which achieved the state-of-the-art performance. However, it is hard to decide which one is the correct target to be generated from the multiple reasonable choices for ReID. It can be seen that the focus of the cross module ReID task is the acquisition equipment, not the difference between the characters and scenes. 

\section{The Proposed Method}
\label{sec:ProposedMethod}
In this section, the proposed method for multiple scenarios in ReID tasks by compositing images is introduced, which is named the Copy and Paste method based on Pose (\textit{CPP}). Taking the cyclist dataset as an example, to construct the cyclist dataset, person datasets and bicycle datasets need to be found. The Market1501 dataset was selected as the pedestrian dataset (\textit{P}) and 710 cyclist images collected by the present authors were used as the cyclist dataset (\textit{R}), in which the mask of each bicycle was obtained. A cyclist from dataset \textit{R} was found and the bicycle thereof was matched with a person in dataset \textit{P}. Notably, the described matching strategy is considerably important, because the effect of the composite image can be directly affected. 

\textbf{Pose Detection.} In order to better match cyclists and pedestrians, the distance from the camera, orientation and posture slope need to be considered. In terms of the distance from the camera, the height difference between two key points can be used for approximation. Because the postures of pedestrians and cyclists in the upper body are similar, the two key points of the middle of the hips and neck were selected as the basis for height discrimination. Said two key points could be also used to calculate the posture slope. The positions of the left and right shoulders could be used to determine the orientation (There are four directions in total, which are up, down, left, and right). At the same time, the position of the middle of the hips could be derived from the left and right positions.

\textbf{Matching Strategy.}
After obtaining the orientation (\textit{T}), height difference (\textit{H}) and posture slope (\textit{K}) of each pedestrian and cyclist, a good matching strategy needed to be built, so as to ensure that the composition image is as realistic as possible, but also has randomness. A prioritization strategy was constructed, with the highest orientation, the second slope, and the lowest height, which ensures that cyclists and pedestrians with the same orientation and posture are combined as much as possible. In order to ensure randomness, the slope and height buckets needed to be considered, and a certain tolerance interval was set ($bin$), which was $bin = 15$.

\textbf{Copy-Paste.}
According to the present matching strategy, for each pedestrian, the cyclist that is the best match can be found. The affine matrix (\textit{M}) can be obtained from the posture slopes of the matched pedestrian and the cyclist and \textit{M}-based affine transformation can be performed on the cyclist for pose correction. Subsequently, the bicycle of the cyclist and the pedestrian can have a relatively balanced slope, before obtaining the image of the bicycle from the mask thereof. Finally, only the pedestrian image and the bicycle need to be copied and pasted to obtain the pedestrian image of the pedestrian riding the bicycle, which can be referred to as the $FakeCyclists$. The specific algorithm pseudo code can be seen in Algorithm \ref{a1}.

\begin{figure}[htb]
\centering
\includegraphics[scale=0.8]{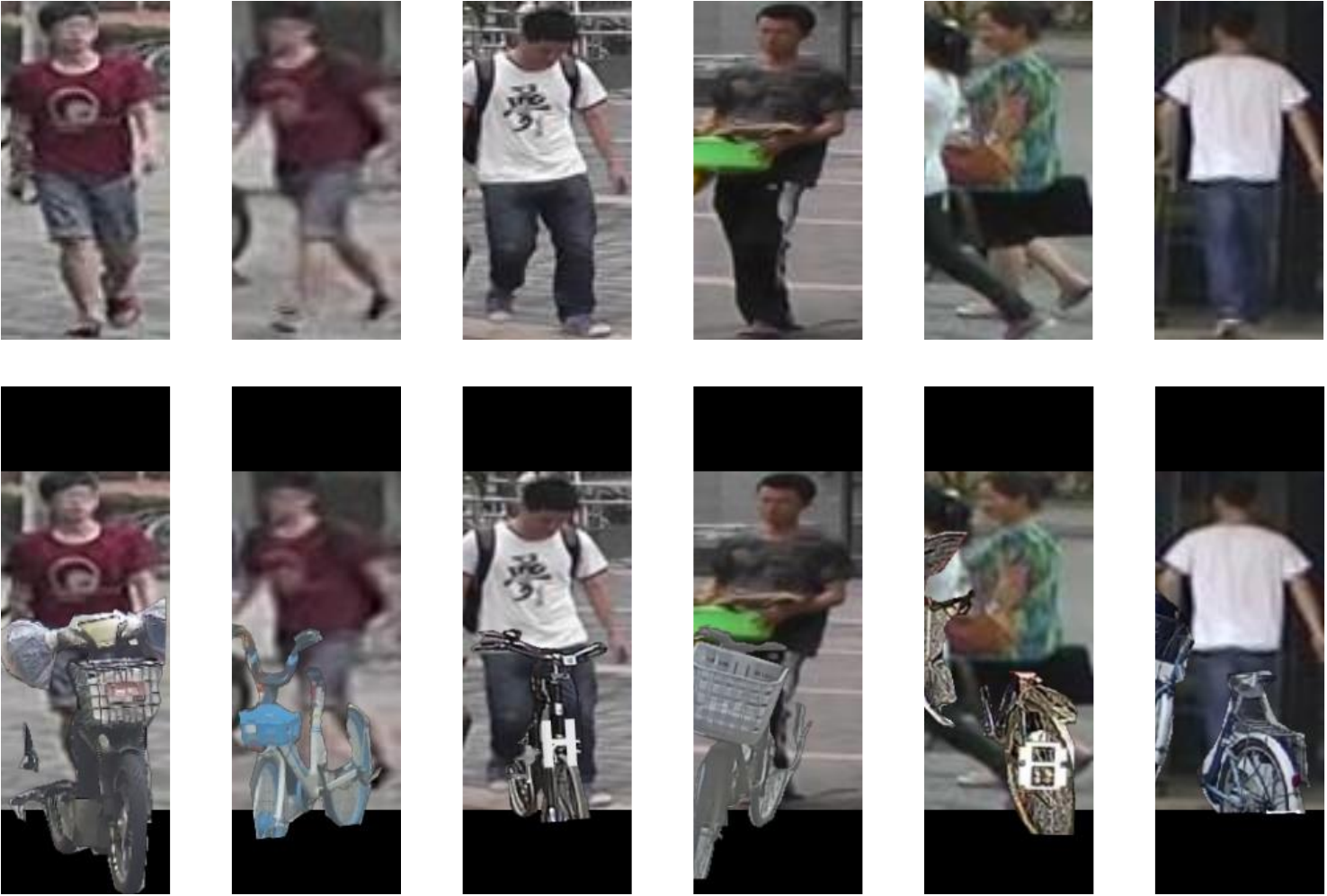}
\quad
\caption{Composition examples. The partially composited cyclist images and pedestrian images. The $1^{st}$ and  $2^{nd}$ columns correspond to the original pedestrian images and the composited cyclist images, respectively.}
\label{f1}
\end{figure}

\begin{algorithm}[H]
\caption{\textbf{The pseudo code of the \textit{CPP}}}
\label{a1}
  \SetKwData{Left}{left}\SetKwData{This}{this}\SetKwData{Up}{up}
  \SetKwFunction{Union}{Union}\SetKwFunction{FindEqualOrientation}{FindEqualOrientation}
  \SetKwFunction{Union}{Union}\SetKwFunction{FindNearestHeight}{FindNearestHeight}
  \SetKwFunction{Union}{Union}\SetKwFunction{FindNearestSlope}{FindNearestSlope}
  \SetKwFunction{Union}{Union}\SetKwFunction{RandomIndex}{RandomIndex}
  \SetKwFunction{Union}{Union}\SetKwFunction{GetAffineMatrix}{GetAffineMatrix}
  \SetKwFunction{Union}{Union}\SetKwFunction{GetAffineImage}{GetAffineImage}
  \SetKwFunction{Union}{Union}\SetKwFunction{CopyAndPaste}{CopyAndPaste}
  \SetKwInOut{Input}{input}\SetKwInOut{Output}{output}
\KwData{the orientation,  height difference and posture slope of all of the pedestrian $T_{P}, H_{P}, K_{P}$, and they is a list. All elements in H and K are multiples of $bin$}
\KwData{the orientation,  height difference and posture slope of all of the cyclist $T_{R}, H_{R}, K_{R}$, and they is a list. All elements in H and K are multiples of $bin$}
\KwResult{Composition images of Cyclists $FakeCyclists$}
\For{$i\leftarrow 0$ \KwTo len($T_{P}$)}{
$T \leftarrow$ \FindEqualOrientation{$T_R$, $T_{P_i}$}\;
$K \leftarrow$ \FindNearestSlope{$K_R$, $K_{P_i}$}\;
$H \leftarrow$ \FindNearestHeight{$H_R$, $H_{P_i}$}\;
$Cyclists \leftarrow$ {$T\cap K \cap H $} \;
$j \leftarrow$ \RandomIndex{$Cyclists$} \;
$AffineMatrix \leftarrow$ \GetAffineMatrix {$K_{P_i}$, $K_{R_j}$} \;
$Bicycle \leftarrow$ \GetAffineImage{$Cyclist$, $AffineMatrix$} $\cap$ {$Bicycle's$ \  $Mask$} \;
$FakeCyclist \leftarrow$ \CopyAndPaste{$P_i, Bicycle$} \;
$FakeCyclists$.append($FakeCyclist$) \;
}
\end{algorithm}
\textbf{Results.} The three person ReID datasets of Market1501, DukeMTMC-reID, and MSMT17 were used to composite corresponding cyclist datasets with 710 cyclist images. For each composite dataset, the identity of the person was unchanged, but the amount of images was doubled. Table \ref{t1} presents information about public person ReID datasets and Figure \ref{f1} shows the partially composited images of cyclist images.

\begin{table}[htb]
\setlength{\abovecaptionskip}{10pt}
\setlength{\belowcaptionskip}{0pt}
\centering
\begin{tabular}{c|c|c|c} 
\hline
Dataset    & Market1501 & DukeMTMC-reID & MSMT17              \\ 
\hline
Identities & 1501       & 1812     & 4101              \\ 
\hline
Cameras    & 6          & 8        & 15                \\ 
\hline
Scene      & outdoor    & outdoor  & outdoor, indoor~  \\
\hline
\end{tabular}
\caption{The brief information of public person ReID datasets}
\label{t1}
\end{table}

\newpage
\section{Network architecture and training}
\label{sec:nsa}
In this section, the network used in the present experiments is described. Inspired by ResNet\cite{r48}, the network could be designed to be considerably deep. The present experiments were based on the FastReID toolbox, which is by far the most general and high-performance toolbox in ReID tasks.

\subsection{FastReID}
FastReID is a research platform that implements state-of-the-art ReID algorithms. In FastReID, the highly modular and extensible design allow for researchers to more easily formulate new research ideas, and the functions of each module in the network are decoupled and split. "Pre-processing" is used to complete image augmentation. "Backbone" is the basic network used to extract image features, such as a ResNet without the last average pooling layer, "Aggregation" is the global pooling layer, which is used to aggregate feature maps generated by backbone into a global feature. "Head" is the part of addressing the global vector generated by the aggregation module, including Batch Normalization (BN) head, Linear head and Reduction head. "Loss" uses the "global vector" generated by "Head" and "Label" and guides the network to better extract key feature from the input image.

\subsection{Component}
"ResNet50" was used as the backbone, and "GeM pooling"\cite{r49} was selected to aggregate feature maps. Then, "BN head"\cite{r50} was selected to balance ID loss and triplet loss\cite{r8}, since the targets of such two losses are inconsistent in the embedding space. Cosine distance is suitable for models optimized by ID loss and Euclidean distance is suitable for triplet loss, relatively. During training, triplet loss\cite{r8} and circle loss\cite{r10} were used to optimize the model.

%
\subsection{Training Strategy}
In order to verify whether the proposed method could be beneficial for image retrieval tasks in general public person ReID datasets, such as Market1501, DukeMTMC-reID, and MSMT17, the composite dataset and original dataset were used to train the model, and compared with the same model trained on the original dataset. The size of input image is $384 \times 128$. 

\textbf{Learning rate.} 
A warm-up method was adopted to improve the over-fitting problem caused by the initial learning rate in the initial stage of model training. A considerably small learning rate, e.g.,$7 \times 10^{-5}$ was set in the initial training phase and then gradually increased during the 4k iterations. Subsequently, the learning rate remained at $7 \times 10^{-4}$ between 4k iterations and 9k iterations. Then, the learning rate started from $7 \times 10^{-4}$ and decayed to $7.7 \times 10^{-7}$ at the cosine rule after 9k iterations. The training finished after 18 iterations.

\textbf{Backbone freeze.} The collected data from the tasks were used to fine-tune the ImageNet pre-trained model. In order to improve the performance of the pre-trained model, the classifier parameters only had to be updated at the beginning training phase (2k iterations). After 2k iterations, all the network parameters for end-to-end training were freed.

\textbf{Aggregation.} The purpose of the aggregation layer is to aggregate features generated by the backbone into global features. Fixed GeM pooling ($p=3$) was used to aggregate features.

\textbf{Head.} "BN head" was used to balance the differences in optimization direction between circle loss and triplet loss due to the different measurement distances. In the test phase, the output of the "BN head" was used as a feature to calculate the distance between the gallery set and the query set.

\textbf{Loss.} During the training phase, triplet loss and circle loss were used to optimize the model. For triplet loss, $P$ identities and $K$ images of each person were selected to constitute a training batch, and the batch size was equal to $B = P \times K$. In this paper, $P = 24, K = 4$ and the $margin=0.3$ were set. For circle loss, $scale = 64, margin = 0.35$ were set. Meanwhile, label smoothing was used to prevent over fitting for a classification task, with $\epsilon = 0.15$.

\section{Experiments and results}
\label{sec:EAR}
\subsection{Ablation study}
The \textit{CPP} method was used to composite images of cyclists in Market1501, DukeMTMC-reID and MTMS17. According to whether the training set, query set and gallery set needed to be composited, 8 datasets were generated for each of the aforementioned sets. Subsequently, the effectiveness of each composite dataset was investigated.

\textbf{Results on Person ReID Datasets.} First, the original training set and the composite training set were used as training sets, and the original query and gallery sets were inferred. The experimental results are shown in Table \ref{t2}. An observation can be made that for the same test sets, the composite training sets exhibited general improvement compared with the original datasets.

\begin{table}[htp]
\setlength{\abovecaptionskip}{10pt}
\setlength{\belowcaptionskip}{0pt}
\centering
\begin{tabular}{c|c|c|cc|cc|cc} 
\hline
\multirow{2}{*}{train} & \multirow{2}{*}{query} & \multirow{2}{*}{gallery} & \multicolumn{2}{c|}{Market1501} & \multicolumn{2}{c|}{DukeMTMC-reID} & \multicolumn{2}{c}{MSMT17}  \\
                       &                        &                          & Rank-1 & mAP                    & Rank-1 & mAP                       & Rank-1 & mAP                \\ 
\hline
w/o                    & w/o                    & w/o                      & 95.69  & 88.74                  & 90.13  & 80.02                     & 81.95  & 57.59              \\ 
\hline
w                      & w/o                    & w/o                      & \textbf{95.97}  & \textbf{89.31}                  & \textbf{90.33}  & \textbf{80.32}                     & \textbf{82.21}  & \textbf{58.26}              \\
\hline
\end{tabular}
\caption{Compared the impact of the two train sets on the results}
\label{t2}
\end{table}

Second, the trained model was used to evaluate the results of different test sets, including whether to use the \textit{CPP} method for different query sets and gallery sets. The experiment results are shown in Table \ref{t3}. An observation can be made that the training set generated by the \textit{CPP} method achieved a better result than that without the \textit{CPP} method.

\raggedbottom
\begin{table}[htp]
\setlength{\abovecaptionskip}{10pt}
\setlength{\belowcaptionskip}{0pt}
\centering
\begin{tabular}{c|c|c|cc|cc|cc} 
\hline
\multirow{2}{*}{train} & \multirow{2}{*}{query} & \multirow{2}{*}{gallery} & \multicolumn{2}{c|}{Market1501} & \multicolumn{2}{c|}{DukeMTMC-reID} & \multicolumn{2}{c}{MSMT17}  \\
                       &                        &                          & Rank-1 & mAP                    & Rank-1 & mAP                       & Rank-1 & mAP                \\ 
\hline
w/o                    & w/o                    & w                        & 95.34  & 63.82                  & 89.14  & 68.59                     & 81.38  & 42.91              \\ 
\hline
w                      & w/o                    & w                        & \textbf{95.40}  & \textbf{85.78}                  & \textbf{89.31}  & \textbf{74.07}                     & \textbf{81.91}  & \textbf{46.21}              \\ 
\hline
w/o                    & w                      & w/o                      & 76.11  & 67.62                  & 82.94  & 70.73                     & 67.94  & 45.23              \\ 
\hline
w                      & w                      & w/o                      & \textbf{94.33}  & \textbf{86.34}                  & \textbf{96.95}  & \textbf{81.32}                     & \textbf{91.32}  & \textbf{57.21}              \\ 
\hline
w/o                    & w                      & w                        & 74.64  & 45.69                  & 81.35  & 59.94                     & 66.24  & 33.44              \\ 
\hline
w                      & w                      & w                        & \textbf{94.37}  & \textbf{83.84}                  & \textbf{85.97}  & \textbf{71.07}                     & \textbf{72.90}  & \textbf{43.24}              \\
\hline
\end{tabular}
\caption{Compared the result of the multiple test set}
\label{t3}
\end{table}

\subsection{Robustness}
In order to evaluate the robustness of the \textit{CPP} method, the public dataset BPReID, which is a public cyclist dataset, was selected to evaluate the performance of the proposed model trained by the \textit{CPP} method. BPReID can be distinguished from existing person ReID datasets in terms of three aspects. First, BPReID was the first bike-person ReID dataset, and has the largest amount of identities by far. Second, BPReID samples from a subset of a real surveillance system, providing a realistic benchmark. Third, there is a long instance between two cameras, providing a wide area benchmark. The experiment results are shown in Table \ref{t4}. An observation can be made that the \textit{CPP} method had good robustness.

\raggedbottom
\begin{table}[htp]
\setlength{\abovecaptionskip}{10pt}
\setlength{\belowcaptionskip}{0pt}
\centering
\begin{tabular}{c|c|c|cc|cc|cc} 
\hline
\multirow{2}{*}{train} & \multirow{2}{*}{query} & \multirow{2}{*}{gallery} & \multicolumn{2}{c|}{Market1501} & \multicolumn{2}{c|}{DukeMTMC-reID} & \multicolumn{2}{c}{MSMT17}  \\
                       &                        &                          & Rank-1 & mAP                    & Rank-1 & mAP                       & Rank-1 & mAP                \\ 
\hline
w/o                    & w/o                    & w                        & 95.34  & 63.82                  & 89.14  & 68.59                     & 81.38  & 42.91              \\ 
\hline
w                      & w/o                    & w                        & \textbf{95.40}  & \textbf{85.78}                  & \textbf{89.31}  & \textbf{74.07}                     & \textbf{81.91}  & \textbf{46.21}              \\ 
\hline
w/o                    & w                      & w/o                      & 76.11  & 67.62                  & 82.94  & 70.73                     & 67.94  & 45.23              \\ 
\hline
w                      & w                      & w/o                      & \textbf{94.33}  & \textbf{86.34}                  & \textbf{96.95}  & \textbf{81.32}                     & \textbf{91.32}  & \textbf{57.21}              \\ 
\hline
w/o                    & w                      & w                        & 74.64  & 45.69                  & 81.35  & 59.94                     & 66.24  & 33.44              \\ 
\hline
w                      & w                      & w                        & \textbf{94.37}  & \textbf{83.84}                  & \textbf{85.97}  & \textbf{71.07}                     & \textbf{72.90}  & \textbf{43.24}              \\
\hline
\end{tabular}
\caption{Shows the robustness of the \textit{CPP}}
\label{t4}
\end{table}

\subsection{Analysis of results}
\begin{figure}[htb]
\centering
\includegraphics[scale=0.6]{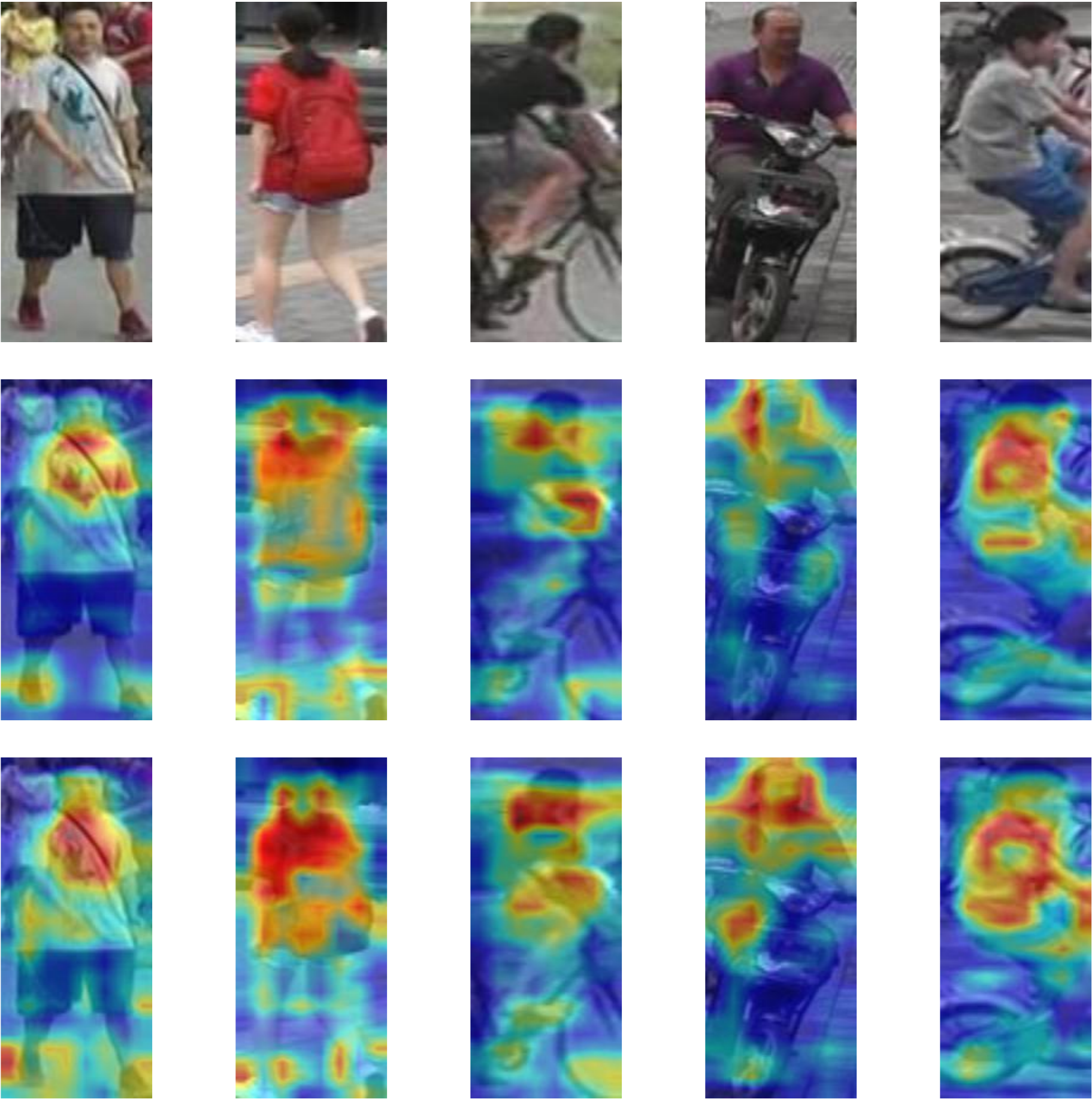}
\quad
\caption{The CAM examples by GradCAMpp\cite{r52, r53}. The position of the model’s attention is reflected in the input image. The $1^{st}$ to $3^{rd}$ columns correspond to the original pedestrian images, the CAM generated without a model using the \textit{CPP} method, and the CAM generated by the model of the \textit{CPP} method, respectively.}
\label{f2}
\end{figure}

The \textit{CPP} method can be considered to involve a more flexible data augmentation. According to the results in Table \ref{t2}, using the \textit{CPP} method on the training set achieved better performance, which could be mainly attributed to the extraction of key features after the \textit{CPP} processing. In order to verify the present conclusions, a Class Activation Map\cite{r51} was visualised on the best model obtained by two methods. The result is shown in Figure \ref{f2}.

\section{Conclusions}
In the present study, new scenarios for ReID tasks were proposed, and the \textit{CPP} method was provided as a solution. Secondly, a scenario based on the mutual retrieval of cyclists and pedestrians was proposed. On the basis of such scenario, the \textit{CPP} method was used to construct corresponding datasets. Finally, the effectiveness and robustness of the \textit{CPP} method on the public person

\bibliography{egbib}
\end{document}